\pdfoutput=1
\documentclass[11pt,letterpaper]{article}
\usepackage{emnlp2017}
\usepackage{times}
\usepackage{graphicx}
\usepackage{stmaryrd}
\usepackage{latexsym}
\usepackage{hyperref}
\usepackage{booktabs}
\usepackage[T1]{fontenc}

\emnlpfinalcopy

\def\emnlppaperid{***}

\title{Ensembling Factored Neural Machine Translation Models for Automatic Post-Editing and Quality Estimation}

\author{Chris Hokamp \\ ADAPT Centre \\ Dublin City University \\
  {\tt chokamp@computing.dcu.ie}}

\date{}

\begin{document}

\maketitle

\begin{abstract}
   This work presents a novel approach to Automatic Post-Editing (APE) and Word-Level Quality Estimation (QE) using ensembles of specialized Neural Machine Translation (NMT) systems. Word-level features that have proven effective for QE are included as input factors, expanding the representation of the original source and the machine translation hypothesis, which are used to generate an automatically post-edited hypothesis. We train a suite of NMT models that use different input representations, but share the same output space. These models are then ensembled together, and tuned for both the APE and the QE task. We thus attempt to connect the state-of-the-art approaches to APE and QE within a single framework. Our models achieve state-of-the-art results in both tasks, with the only difference in the tuning step which learns weights for each component of the ensemble. 
\end{abstract}

\section{Introduction}

Translation destined for human consumption often must pass through multiple editing stages. In one common scenario, human translators correct machine translation (MT) output, correcting errors and omissions until a perfect translation has been produced. Several studies has shown that this process, referred to as "post-editing", is faster than translation from scratch~\cite{Specia2011}, or interactive machine translation~\cite{Green:2013}. 

A relatively recent line of research has tried to build models which correct errors in MT automatically~\cite{Simard2007,bojar-EtAl:2015:WMT,junczysdowmunt-grundkiewicz:2016:WMT}. Automatic Post-Editing (APE) typically views the system that produced the original translation as a black box, which cannot be modified or inspected. An APE system has access to the same data that a human translator would see: a source sentence and a translation hypothesis. The job of the system is to output a corrected hypothesis, attempting to fix errors made by the original translation system. This can be viewed as a sequence-to-sequence task~\cite{Sutskever:2014:SSL:2969033.2969173}, and is also similar to multi-source machine translation~\cite{ZophK16,conf/naacl/FiratCB16}. However, APE intuitively tries to make the minimum number of edits required to transform the hypothesis into a satisfactory translation, because we would like our system to mimic human translators in attempting to minimize the time spent correcting each MT output. This additional constraint on APE models differentiates the task from multi-source MT.

The Word Level QE task is ostensibly a simpler version of APE, where a system must only decide whether or not each word in an MT hypothesis belongs in the post-edited version -- it is not necessary to propose a fix for errors. Most recent work has considered word-level QE to be a sequence labeling task, and employed the standard tools of structured prediction to solve it, i.e. structured predictors such as CRFs or structured SVMs, which take advantage of sparse representations and very large feature sets, as well as dependencies between labels in the output sequence~\cite{LogachevaHS16,martins-EtAl:2016:WMT}. However, Martins et al.~\shortcite{martinspushing} recently proposed a new method of word-level QE using APE, which simply uses an APE system to produce a "pseudo-post-edit" given a source sentence and an MT hypothesis. Their approach, which we call \textbf{APE-QE}, is the basis of the work presented here. In APE-QE, the original MT hypothesis is then aligned with the pseudo-post-edit from the APE system using word level edit-distance, and words which correspond to \emph{Insert} or \emph{Delete} operations are labeled as incorrect. Note that this also corresponds exactly to the way QE datasets are currently created, with the only difference being that human post-edits are typically used to create gold-standard data~\cite{bojar-EtAl:2015:WMT}. 

A key similarity between the QE and APE tasks is that both use information from two sequences: (1) the original source input, and (2) an MT hypothesis. Martins et al.~\shortcite{martinspushing}, showed that APE systems with no knowledge about the QE task already provide a very strong baseline for QE. Because the essential training data for the APE and QE tasks is identical, consisting of parallel triples of $ (SRC, MT, PE) $, it is also natural to consider these tasks as two subtasks that make use of a single underlying model. 

In this work, we explicitly design ensembles of NMT models for both word-level QE, and APE. This approach builds upon the approach presented in Martins et al.~\shortcite{martinspushing}, by incorporating features which have proven effective for Word Level QE as "factors" in the input to Neural Machine Translation (NMT) systems. We achieve state-of-the-art results in both Automatic Post-Editing and Word-Level Quality Estimation, matching the performance of much more complex QE systems, and significantly outperforming the current state-of-the-art in APE. 

\vspace{0.2cm}
\noindent The main contributions of this work are:

\begin{itemize}
    \item Novel Input Representations for Neural APE models
    \item New tuned ensembles for APE-QE
    \item An open-source decoder supporting ensembles of models with different inputs\footnote{code avaiable at \url{https://github.com/chrishokamp/constrained_decoding}}
\end{itemize}

The following sections discuss our approach to creating hybrid models for APE-QE, which should be able to solve both tasks with minimal modification. 

\section{Related Work}

Two important lines of research have recently made breakthroughs in QE and APE.

\subsection{Automatic Post-Editing}

APE and QE training datasets consist of $ (SRC, MT, PE) $ triples, where the post-edited reference is created by a human translator in the workflow described above. However, publicly available APE datasets are relatively small in comparison to parallel datasets used to train machine translation systems. Junczys-Dowmunt and Grundkiewicz~\shortcite{junczysdowmunt-grundkiewicz:2016:WMT} introduce a method for generating a large synthetic training dataset from a parallel corpus of $ (SRC, REF) $ by first translating the reference to the source language, and then translating this "pseudo-source" back into the target language, resulting in a ``pseudo-hypothesis" which is likely to be more similar to the reference than a direct translation from source to target. The release of this synthetic training data was a major contribution towards improving APE. 

Junczys-Dowmunt and Grundkiewicz~\shortcite{junczysdowmunt-grundkiewicz:2016:WMT} also present a framework for ensembling SRC $\rightarrow $ PE and SRC $\rightarrow $ PE NMT models together, and tuning for APE performance. Our work extends this idea with several new input representations, which are inspired by the goal of solving both QE and APE with the same model.


\subsection{Quality Estimation}


Martins et al.~\shortcite{martins-EtAl:2016:WMT} introduced a stacked architecture, using a very large feature set within a structured prediction framework to achieve a large jump in the state of the art for Word-Level QE. Some features are actually the outputs of standalone feedforward and recurrent neural network models, which are then stacked into the final system. Although their approach creates a very good final model, the training and feature extraction steps are quite complicated. An additional disadvantage of this approach is that it requires "jackknifing" the training data for the standalone models that provide features to the stacked model, in order to avoid overfitting in the stacked ensemble. This requires training $ k $ versions of each model type, where $ k $ is the number of jackknife splits.

Our approach is most similar to Martins et al.~\shortcite{martinspushing}, the major differences are: we do not use any internal features from the original MT system, and we do not need to "jackknife" in order to create a stacked ensemble. Using only NMT with attention, we are able to surpass the state-of-the-art in APE and match it in QE. 

\subsection{Factored Inputs}
Alexandrescu and Kirchoff~\shortcite{Alexandrescu:2006:NLM} introduced linguistic factors for neural language models. The core idea is to learn embeddings for linguistic features such as part-of-speech (POS) tags and dependency labels, augmenting the word embeddings of the input with additional features. Recent work has shown that NMT performance can also be improved by concatenating embeddings for additional word-level "factors" to source-word input embeddings~\cite{SennrichH16:factors}. The input representation $ e_{j} $ for each source input $ x_{j} $ with factors $ F $ thus becomes Eq.~\ref{eq:factored_input}:

\begin{equation}
    e_{j} = \bigparallel_{k=1}^{|F|} \mathbf{E}_{k} x_{jk} 
    \label{eq:factored_input}
\end{equation}

\noindent where $ \bigparallel $ indicates vector concatenation, $ \mathbf{E}_{k} $ is the embedding matrix of factor $ k $, and $ x_{jk} $ is a one hot vector for the $k$-th input factor.
 


\begin{table*}[ht!]
\begin{center}
\begin{tabular}{p{0.28\linewidth}p{0.15\linewidth}p{0.15\linewidth}p{0.15\linewidth}p{0.15\linewidth}}
\multicolumn{5}{c}{\textbf{WMT 2016 Dev}} \\
\hline
\bf Model Input           & \bf BLEU          & \bf TER $\downarrow$            & \bf F1-Mult   & \bf Accuracy             \\
\hline 
\bf WMT 16 Best           & 68.94             &   .215             & .493          &  --                    \\
\bf Martins et al (2017)  & --                &   --               & \bf{.568}          &  --                    \\
\bf SRC                   & 55.47             &   .315             & .506          &  .803              \\
\bf MT                    & 66.66             &   .232             & .328          &  .834          \\
\hline
\bf MT-aligned            & 68.32             &   .215             & .437          &  .852                 \\
\bf SRC+MT                & 69.17             &   .211             & .477          &  .857                 \\
\bf SRC+MT-factor         & 69.75             &   .209             & .484          &  .859                 \\ 
\bf Avg-All Baseline      & 71.02             &   .199             & .476          &  .862               \\ 
\bf Avg-All APE-Tune      & \bf 71.22        &   \bf .197          & .510          &  \bf .866                 \\ 
\bf Avg-All QE-Tune       & 66.92             &   .228             & .554          &  .857                \\ 
\bf 4-SRC+Avg-All QE-Tune & 67.16             &   .225             & .567          &  .860                \\ 
\hline
\vspace{0.25cm} \\
\multicolumn{5}{c}{\textbf{WMT 2016 Test}} \\
\hline
\bf Model Input           & \bf BLEU          & \bf TER $\downarrow$            & \bf F1-Mult   & \bf Accuracy             \\
\hline 
\bf WMT Baseline          & 62.11             &   .248             & .324          &   --     \\
\bf WMT 16 Best           & 67.65             &   .215             & .493          &   --                     \\
\bf Martins et al (2017)  & 67.62             &   .211             & \bf .575      &   --                    \\
\bf SRC                   & 55.58             &   .304             & .519          &  .809              \\
\bf MT                    & 65.85             &   .234             & .347          &  .837         \\
\hline
\bf MT-aligned            & 67.69             &   .216             & .447          &  .854                 \\
\bf SRC+MT                & 68.03             &   .212             & .477          &  .857                 \\
\bf SRC+MT-factor         & 68.28             &   .211             & .473          &  .857                  \\ 
\bf Avg-All Baseline      & \bf 70.05         &   .198             & .492          &  .865              \\ 
\bf Avg-All APE-Tuned     & 70.04             &   \bf .196         & .516          &  \bf .868               \\ 
\bf Avg-All QE-Tuned      & 66.93             &   .219             & .573          &  .864               \\ 
\bf 4-SRC+Avg-All QE-Tune & 66.94             &   .219             & \bf .575      &  .865                \\ 
\hline
\end{tabular}
\end{center}
\caption{\label{table:wmt16_results} Results for all models and ensembles on WMT 16 development and test datasets}
\end{table*}

\section{Models}

\begin{figure}[!t]  
\hspace*{-1cm}
\centering
\includegraphics[width=0.42\textwidth]{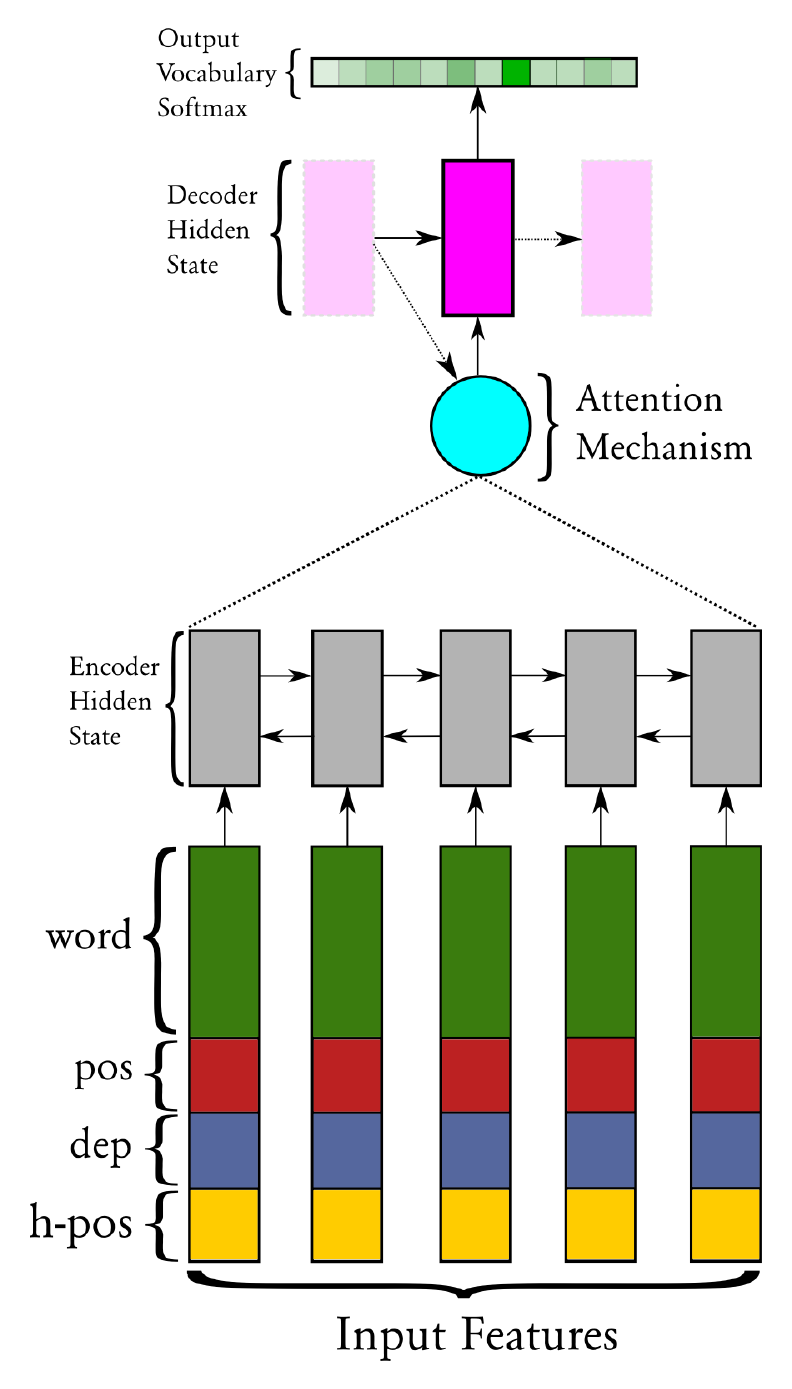}  
\vspace{-0.25cm}
\caption{Schematic of the architecture of our factored NMT systems}
\label{fig:factored_input}
\end{figure}

In this section we describe the five model types used for APE-QE, as well as the ensembles of these models which turn out to be the best-performing overall. We design several features to be included as inputs to APE. The operating hypothesis is that that features which haven proven useful for Quality Estimation should also have a positive impact upon APE performance. 

Our baseline models are the same models used in Junczys-Dowmunt~\shortcite{junczysdowmunt-grundkiewicz:2016:WMT}\footnote{These models have been made available by the authors at \url{https://amunmt.github.io/examples/postedit/}}. The authors provide trained $ SRC \rightarrow PE $ and $ MT \rightarrow PE $ models, which correspond to the last four checkpoints from fine-tuning the models on the \emph{500K} training data concatenated with the task internal APE data upsampled 20 times. These models are referred to as \textbf{SRC} and \textbf{MT}. 

\subsection{Word Alignments}

Previous work has shown that alignment information between source and target is a critical component of current state-of-the-art word level QE systems~\citep{kreutzer-schamoni-riezler:2015:WMT,martins-EtAl:2016:WMT}. The sequential inputs for structured prediction, as well as the feedforward and recurrent models in existing work obtain the source-side features for each target word using the word-alignments provided by the WMT task organizers. However, this information is not likely to be available in many real-world usecases for Quality Estimation, and the use of this information also means that the MT system used to produce the hypotheses is not actually a "black box", which is part of the definition of the QE task. Clearly, access to the word-alignment information of an SMT system provides a lot of insight into the underlying model.

Because our models rely upon synthetic training data, and because we wish to view the MT system as a true black-box, we instead use the \textbf{SRC} NMT system to obtain these alignments. The attention model for NMT produces a normalized vector of weights at each timestep, where the weights can be viewed as the "alignment probabilities" for each source word~\cite{bahdanau2014neural}. In order to obtain the input representation shown in table~\ref{tab:input_examples}, we use the source word with the highest weight from the attention model as an additional factor in the input to another MT-aligned $ \rightarrow $ PE system. The \textbf{MT-aligned} $ \rightarrow $ PE system thus depends upon the \textbf{SRC} $ \rightarrow $ PE system to produce the additional alignment factor. 

\subsection{Inputting Both Source and Target}

Following Crego et al.~\shortcite{crego2016systran}, we train a model which takes the concatenated source and MT as input. The two sequences are separated by a special \emph{BREAK} token. We refer to this system as \textbf{SRC+MT}.

\subsection{Part-of-Speech and Dependency Labels}

Sennrich and Haddow~\shortcite{SennrichH16:factors} showed that information such as POS tags, NER labels, and syntactic roles can be included in the input to NMT models, generally improving performance. Inspired by this idea, we select some of the top performing features from Martins et al.~\cite{martins-EtAl:2016:WMT}, and include them as input factors to the \textbf{SRC+MT-factor} model. The base representation is the concatenated SRC+MT (again with a special \emph{BREAK} token). For each word in the English source and the German hypothesis, we obtain the part-of-speech tag, the dependency relation, and the part-of-speech of the head word, and include these as input factors. For both English and German, we use spaCy\footnote{https://spacy.io/} to extract these features for all training, development, and test data. The resulting model is illustrated in figure~\ref{fig:factored_input}.

\subsection{Extending Factors to Subword Encoding}

Our NMT models use subword encoding~\cite{SennrichSubword}, but the additional factors are computed at the word level. Therefore, the factors must also be segmented to match the BPE segmentation. We use the \{BILOU\}- prefixes common in sequence-labeling tasks such as NER to extend factor vocabularies and map each word-level factor to the subword segmentation of the source or target text.

Table~\ref{tab:input_examples} shows the input representations for each of the model types using an example from the WMT 2016 test data. 

\subsection{Ensembling NMT Models}

We average the parameters of the four best checkpoints of each model type, and create an ensemble of the resulting five models, called \textbf{Avg-All Baseline}. We then tune this ensemble for TER (APE) and F1-Mult (QE), using MERT~\cite{Och:2003:MER:1075096.1075117}. The tuned models are called \textbf{Avg-All APE-Tuned} and \textbf{Avg-All QE-Tuned}, respectively. After observing that source-only models have the best single-model QE performance (see section~\ref{sec:results}), we created a final F1-Mult tuned ensemble, consisting of the four individual \textbf{SRC} models, and the averaged models from each other type (an ensemble of eight models total), called \textbf{4-SRC+Avg-All QE-Tune}.

\subsection{Tuning}

Table~\ref{table:tuning_weights} shows the final weights for each ensemble type after tuning. In line with the two-model ensemble presented in Martins et al.~\shortcite{martinspushing}, tuning models for F1-Mult results in much more weight being allocated to the SRC model, while TER tuning favors models with access to the MT hypothesis.  



\begin{table}[ht!]
\begin{center}
\begin{tabular}{r c c}
               & APE (TER) & QE (F1-Mult) \\
SRC            & .162      & ~.228  \\
MT             & .003      & -.183 \\
MT-aligned     & .203      & ~.229  \\
SRC+MT         & .222      & ~.231  \\
SRC+MT-factor  & .410      & ~.129  \\
\end{tabular}
\end{center}
\caption{\label{table:tuning_weights} Final weights for each model type after 10 iterations of MERT for tuning objectives TER and F1-Mult.}
\end{table}


\begin{table*}[t]
\small
\centering
\begin{tabular}{@{\hspace{0cm}}l@{\hspace{.2cm}}p{0.82\textwidth}@{\hspace{0cm}}}
\toprule
SRC & auto vector masks apply predefined patterns as vector masks to bitmap and vector objects . \\
\vspace{.2cm}

MT & automatische Vektor- masken vordefinierten Mustern wie Vektor- masken , Bitmaps und Vektor- objekte anwenden . \\
\vspace{.2cm}

MT-aligned & automatische|auto Vektor-|vector masken|masks vordefinierten|apply Mustern|patterns wie|as Vektor-|vector masken|masks ,|to Bitmaps|to und|and Vektor-|vector objekte|objects anwenden|apply .|. \\
\vspace{.2cm}

SRC+MT & auto vector masks apply predefined patterns as vector masks to bitmap and vector objects . BREAK automatische Vektor- masken vordefinierten Mustern wie Vektor- masken , Bitmaps und Vektor- objekte anwenden . \\
\vspace{.2cm}

SRC+MT Factored & Auto|JJ|amod|NNS vector|NN|compound|NNS masks|NNS|nsubj|VBP apply|VBP|ROOT|VBP predefined|VBN|amod|NNS patterns|NNS|dobj|VBP as|IN|prep|NNS vector|NN|compound|NNS masks|NNS|pobj|IN to|TO|aux|VB bitmap|VB|relcl|NNS and|CC|cc|VB vector|NN|compound|NNS objects|NNS|conj|VB .|.|punct|VBP BREAK|BREAK|BREAK|BREAK Automatische|ADJA|nk|NN Vektor-|B-NN|B-sb|B-VVINF masken|I-NN|I-sb|I-VVINF vordefinierten|ADJA|nk|NN Mustern|NN|pd|NN wie|KOKOM|cd|NN Vektor-|B-NN|B-cj|B-KOKOM masken|I-NN|I-cj|I-KOKOM ,|\$,|punct|NN Bitmaps|NN|cj|NN und|KON|cd|NN Vektor-|B-NN|B-cj|B-KON objekte|I-NN|I-cj|I-KON anwenden|VVINF|ROOT|VVINF .|\$.|punct|VVINF \\

\midrule

PE (Reference) & Automatische Vektormasken wenden vordefinierte Mustern als Vektormasken auf Bitmap- und Vektorobjekte an . \\

Gold Tags & OK OK BAD OK BAD OK BAD BAD OK OK BAD OK \\

\bottomrule
\end{tabular}
\caption{\label{tab:input_examples} Examples of the input for the five model types used in the APE and QE ensembles. The pipe symbol `|' separates each factor. `-' followed by whitespace indicates segmentation according to the subword encoding.}
\end{table*}

\section{Experiments}

\begin{table}[h!]
\small
\begin{center}
\begin{tabular}{r c c c}
\textbf{Model} & \textbf{General} & \textbf{Fine-tune} & \textbf{Min-Risk} \\
\hline
MT-aligned   & 60.31  & 67.54  & --       \\
SRC+MT       & 59.52  & 68.68  & 69.44    \\
SRC+MT-factor& 57.59  & 68.26  & 69.76    \\
\end{tabular}
\end{center}
\caption{\label{table:training_dev_results} Best BLEU score on dev set after each of the training stages. \textit{General} is training with 4M instances, \textit{Fine-tune} is training with 500K + upsampled in-domain data, \textit{Min-Risk} uses the same dataset as \textit{Fine-tune}, but uses a minimum-risk loss with BLEU score as the target metric.}
\end{table}

All of our models are trained using Nematus~\cite{sennrich-EtAl:2017:EACLDemo}. At inference time we use our own decoder, which supports weighted log-linear ensembles of Nematus models\footnote{\url{https://github.com/chrishokamp/constrained_decoding}}. Following Junczys-Dowmunt and  Grundkiewicz~\shortcite{junczysdowmunt-grundkiewicz:2016:WMT}, we first train each model type on the large (4M) synthetic training data, then fine tune using the 500K dataset, concatenated with the task-internal training data upsampled 20x. Finally, for \textbf{SRC+MT} and \textbf{SRC+MT-factor} we continued fine-tuning each model for a small number of iterations using the min-risk training implementation available in Nematus~\cite{DBLP:conf/acl/ShenCHHWSL16}. Table~\ref{table:training_dev_results} shows the best dev result after each stage of training. 

For both APE and QE, we use only the task-specific training data provided for the WMT 2017 APE task, including the extra synthetic training data\footnote{\url{http://www.statmt.org/wmt17/ape-task.html}}. However, note that the SpaCy models used to extract features for the factored models are trained with external data -- we only use the off-the-shelf models provided by the SpaCy developers.

To convert the output sequence from an APE system into $ {OK, BAD} $ labels for QE, we use the APE hypothesis as a "pseudo-reference", which is then aligned with the original MT hypothesis using TER~\citep{Snover06astudy}. 



\section{Results}
\label{sec:results}

Table~\ref{table:wmt16_results} shows the results of our experiments using the WMT 16 development and test sets. For each system, we measure performance on BLEU and TER, which are the metrics used in APE task, and also on F1-Mult, which is the primary metric used for the Word Level QE task. Overall tagging accuracy is included as a secondary metric for QE. 

All systems with input factors significantly improve APE performance over the baselines. For QE, the trends are less clear, but point to a key difference between optimizing for TER vs. F1\_product: F1\_product optimization probably lowers the threshold for "changing" a word, as opposed to copying it from the MT hypothesis. This hypothesis is supported by the observation that the source-only APE system outperforms all other single models on the QE metrics. Because the source-only systems cannot resort to copying words from the input, they are forced to make the best guess about the final output, and words which are more likely to be wrong are less likely to be present in the output. If input factors were used with a source-only APE system, the performance on word-level QE could likely be further improved. However, this hypothesis needs more analysis and experimentation to confirm.

\section{Conclusion}
\label{sec:conclusion}

This work has presented APE-QE, unifying models for APE and word-level QE by leveraging the flexibility of NMT to take advantage of informative features from QE. Models with different input representations are ensembled together and tuned for either APE or QE, achieving state of the art performance in both tasks. The complementary nature of these tasks points to future avenues of exploration, such as joint training using both QE labels and reference translations, as well as the incorporation of other features as input factors. 


\section*{Acknowledgments}
This project has received funding from Science Foundation Ireland in the ADAPT Centre for Digital Content Technology (www.adaptcentre.ie) at Dublin City University funded under the SFI Research Centres Programme (Grant 13/RC/2106) co-funded under the European Regional Development Fund and the European Union Horizon 2020 research and innovation programme under grant agreement 645452 (QT21). Marcin Junczys-Dowmunt provided essential guidance on the tuning implementation for APE and QE ensembles. 

\bibliography{emnlp2017}
\bibliographystyle{emnlp_natbib}

\end{document}